# Abusive Language Detection and Characterization of Twitter Behavior

## Dincy Davis[1*], Reena Murali[2], Remesh Babu K.R.[3]

[1,2]Dept. of Computer Science and Engineering, Government Engineering College, Palakkad, India
[3]Dept. of Information Technology, Government Engineering College, Palakkad, India

*Corresponding Author: dincydavis17@gmail.com, Tel.: +91-9495057771

***Abstract—*** Abusive language refers to an insult or vulgarity which harass or deceive the target. Social media is a famous platform for the people to express their opinions publicly and to interact with other people in the world. Some of them may misuse their freedom of speech to bully others through abusive language. This will leads to the need for detecting abusive speech. Otherwise, it may severely impact the user's online experience. It may be a time-consuming task if the detection and removal of such offensive material are done manually. Also, human supervision is unable to deal with large quantities of data. Therefore automatic abusive speech detection has become essential to be addressed effectively. For detecting abusive speech, context accompanying abusive speech is very useful. In this work, abusive language detection in online content is performed using Bidirectional Recurrent Neural Network (BiRNN) method. Here the main objective is to focus on various forms of abusive behaviors on Twitter and to detect whether a speech is abusive or not. The results are compared for various abusive behaviors in social media, with Convolutional Neural Netwrok (CNN) and Recurrent Neural Network (RNN) methods and proved that the proposed BiRNN is a better deep learning model for automatic abusive speech detection.

## I. INTRODUCTION

The Internet has provided an opportunity for the world wide users to meet and express their opinions, resulting in a massive amount of user-generated data in online platforms. These opportunities also apply to those with malicious intentions, who can effortlessly and anonymously express hateful statements to large groups or targeting specific individuals. Identifying hate speech is a matter of priority for sites that allow user-generated content. Although there is no formal definition, abusive speech is commonly defined as a strongly rude, impolite and hurtful language targets specific group characteristics, such as ethnicity, religion or gender. The large and increasing amount of user-generated data on social media makes it very difficult to identify and eliminate online hateful speech, which motivates research on how advanced technologies can help solve the issue. This study will focus on how the technologies can help to automatically detect abusive posts by exploring information beyond the actual textual content.

Also there are various forms of abusive behavior exist in social media. Founta et al. [1] have build a data set of robust collection of abuse related labels in order to characterize the abuse related tweets. Those were used crowd sourcing annotation and the labels include abusive and aggressive, hateful, offensive, cyber-bullying, etc to distinguish between various expressions of online abuse.

An example for hateful tweet that have been posted in Twitter is shown in Figure 1.

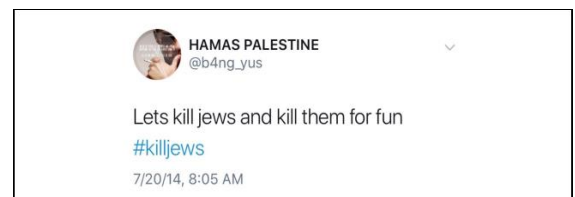

Figure 1. Example of Abusive language

A large number of studies has been done in recent years to develop automatic methods for the detection of abusive languages in social media platforms. These typically employed methods depend on Natural Language Processing (NLP) and Machine Learning (ML) methods, which involve classification of textual content as normal or abusive. Although current methods have reported promising results, their evaluations are largely biased towards the prediction of normal content, instead of the detection and classification of exact abusive content. These results suggest that it is much harder to detect abusive content than is normal. However, it is more important to be able to correctly detect and identify specific types of hate speech. Since deep learning methods has considerable usefulness in recent years and studies showed that RNN provided best results than other neural networks for abusive comment analysis, the work proposed here using BiRNN which is an extension of RNN.

Motivated by these observations, this paper proposes a BiRNN structures that are empirically shown to be very effective feature extractors for identifying the specific types of abusive language by using a multi labeled dataset. The remaining part of this paper is organized as follows: Section II contains the related work of abusive language detection; Section III discusses about the system architecture, deep learning model and dataset used in this work; Section IV presents the implementation phases, results, performance analysis and performance comparison of proposed work with existing techniques. Section V concludes research work with future directions.

## II.   RELATED WORK

This section deals with the previous research works for abusive language detection.

*A. Methodology-1 Convolutional Neural Networks (CNN)*
Badjatiya et al. [2] were one of the first to apply deep learning architectures for the task of hate speech detection. They used a Gradient-Boosted Decision Tree (GDBT) on word embedding learned using a CNN. Embedding learned from deep neural network models when combined with gradient boosted decision trees led to best accuracy values. They found them to significantly outperform the existing methods.

Gambäck et al. in his methodology [3] used CNN for Twitter hate-speech text classification. The classifier assigns each tweet into predefined four categories: racism, sexism, both (racism and sexism) and non-hate-speech. Four CNN models for character 4-grams, word vectors based on semantic information built using word2vec, randomly generated word vectors, and word vectors combined with character n-grams were trained in this work and the model based on word2vec embedding performed best.

The following two-step process was used in the work of Park et al. [4]. They first perform classification on abusive language, such that, detect whether or not a tweet was abusive, and then use another classifier to classify it into specific classes as racist or gender-based. They were using a model of Hybrid CNN which is a variant of CNN that uses both words and characters to make classification decision. This experiment has shown that it can boost the performance of models such as logistic regression.

*B. Methodology-2 Recurrent Neural Network (RNN)*
There are several architectures that uses variant RNN. Pavlopoulos et al. [5] proposes the use of RNN with improved accuracy on toxicity and personal attack dataset. In addition, another dataset was released, with 1.6 million manually annotated user comments from the Greek Sports Portal (Gazzetta) and embedding trained on this data. This work has shown that the Gated Recurrent Neural Network (GRU) RNN operating on word embedding outperforms the previous state of the art that used the Logistic Regression or Multi Layer Perceptron (MLP) Classifier with character or word n-gram features.

D. Kumar et al. [6] introduced the use of BiRNN for abusive language detection, but it is limited to the use of multiple abuse related classification. The paper [7] conducted a comparative analysis of different learning models on the Hate and Abusive Speech dataset on Twitter. Their experimental results show that GRU RNN have better accuracy up to 0.801 F1. Similarly in paper [8] proved that Long Short Term Memory (LSTM) RNN is better for sequence learning problems than standard CNN.

## III.   METHODOLOGY

This section gives the system architecture and the details of the dataset and deep learning model employed in this work.

*A. Dataset*
The lack of a benchmark dataset for the task of hate speech detection is an issue as it becomes difficult to compare methods and results that are based on different data and annotations. In addition, the datasets are created for different tasks, and therefore have different characteristics and display different types of hate speech. Creating datasets for this task is time consuming, as the number of hateful instances in online communities is relatively few, but it is necessary to have a representable amount of such instances in a dataset. There are also several datasets that have not been made publicly available. This may be due to privacy issues or considering the content of the datasets, that is, the profanity and offensive language. In this work, using a dataset implemented by Founta et al. [1] which have four classes as abusive, hateful, spam and normal. As mentioned before, this proposed system aims at developing a classifier that classifies text for a given user comment by using BiRNN. This dataset contains 100k tweeter posts and the distribution is shown in Table 1.

Table 1: Dataset summary

| Labels | Normal | Spam | Hateful | Abusive |
|---|---|---|---|---|
| Number | 42,932 | 9,757 | 3,100 | 15,115 |
| Percentage(%) | 60.5 | 13.8 | 4.4 | 21.3 |

*B. BiRNN*
BiRNN connect two hidden layers of different directions to the same output. With this form of conceptual deep learning, the output layer can simultaneously obtain information from past (backward) and future (forward) states. This was developed to increase the amount of input available information to the network. For example, MLP have limitations on the flexibility of input data, as they demand their input data to be fixed. Standard RNN also have limitations, as future input information can not be reached from the current state. BiRNN are particularly useful when an input context is needed. For example, in the case of handwriting recognition, awareness of the letters before and after the current letter can be improved.

Figure 3 shows the general BiRNN model [9] which involves a text input of embedding followed by a hidden layer. The softmax function produce a vector that represents the probability of a list of classes and uses these to generate most probable class in output layer.

In short, the network architecture of our model has the following structure:
- Embedding Layer : This layer converts each word into an embedded vector.
- Hidden Layer: The hidden layer is a BiRNN. The output of this layer is a fixed size interpretation of its input.
- Output Layer: In the output layer, the interpretation learned from the output layer, RNN passes through a fully connected neural network with a softmax output node which classifies the text as abusive, hateful, spam or normal.

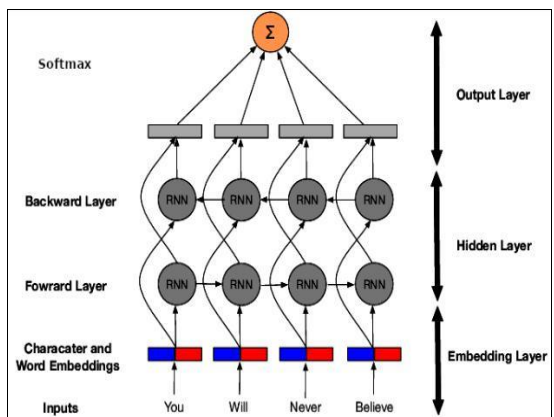
Figure 3.  General BiRNN architecture

Glove approach [10] is used for the embedding of words. These embedded words are then passed to the deep learning module. Table 2 shows the implementation details of the deep learning model. The proposed deep learning model is implemented using tensorflow. Numpy python library is used in this phase to format the input as an array.

Table 2: BiRNN model specifications

| Parameter | Value |
|---|---|
| Layers | Input, Embedding, LSTM (Hidden layers), Dense |
| Embedding size dimension | 300 |
| Size of word dictionary | 137898 |
| Size of dataset | 61557 |
| Activation | Softmax |
| Optimizer | Adam |
| Loss | categorical cross entropy |
| Epochs | 200 |
| Batch Size | 64 |

*C. System Architecture*
The proposed system classifies comments from Twitter platform and the classes includes robust set of abuse related labels. This work developed a BiRNN detection model by training the architecture with Twitter data. Figure 2 shows the basic architecture of the proposed abusive language detection system.

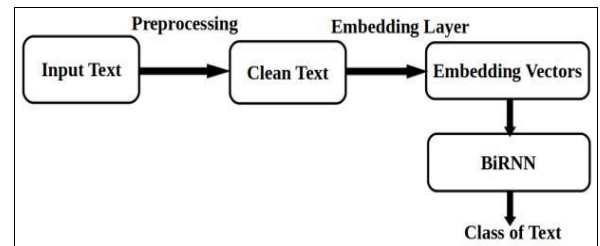
Figure 2.  System Architecture

The overall system architecture can be considered as three main modules namely:
1. Text pre-processing module
2. Embedding module
3. Model generation

*1. Text Pre-processing Module*
Preprocessing can be considered as a step for removing noise, and is done for a more easily extraction of features and information. When it comes to online writing, people often disregard grammar rules, misspell words, and use abbreviations. For a machine to be able to understand and make sense of such human-written language, the preprocessing step becomes essential. Texts are preprocessed to remove any content that does not have any useful information. Since social media content is noisy and may contain typos, short hands and intentional obfuscations, preprocessing is an essential step when we are dealing with user-generated data that have been posted online. The workflow of preprocessing module include:
- Load the dataset using pandas.
- Split the corpus into training set and test set.
- Apply a preprocessing transform to the input sets which contains various user comments and its corresponding classes and also the same preprocessing task are carried out when a particular input text is given by the user.

The sub tasks involved in the text preprocessing are as shown in Figure 4.
The chosen dataset contains 100k tweets. In the proposed method first it removes all the unwanted features such as excess white spaces, spaces at the beginning, etc. Then it converts all the characters present in the input text to lowercase. It is one of the simpler and effective text processing task. Lower casing actually help in cases where our dataset is not very large and it also significantly helps with the consistency of expected output. Also it makes easier to compare words and to reduce the sample size. Tokenization is a process of converting the text in to tokens that retain all the essential information about the data. Next we are replacing the elongated text as short by removing repeated characters. Then we are doing word segmentation in order to deal with the concatenated words like 'stupidperson' or 'stupid person' and at last we are passing these text in to spelling correction mechanism to ensure the correctness of text.

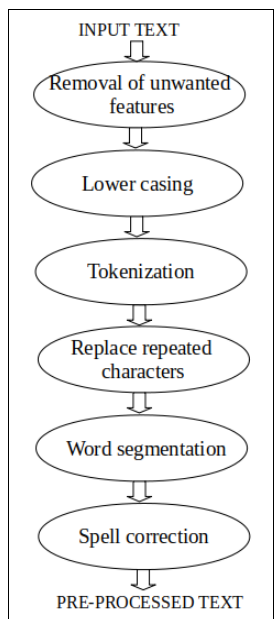

Figure 4. Sub tasks of text preprocessing

As mentioned before, we don't need all the information from the corpus. Thus we filter only necessary information from it, and remove the null spaces as well.

*2. Embedding Module*
Embedding vector generation is the next task that we need to consider. After cleaning the data, It should be converted in to such a way a machine can understand. Word embedding can be considered both as word representations and features. As features, embedding may replace the presence or frequency of particular words, where words that are used similarly also have similar representations. For that purpose we use this embedding layer. Different type of pretrained embedding are there. Examples of models for constructing word embedding from text include word2vec and GloVe, and these will be further explained due to their popularity. Word2vec is a predictive algorithm developed by Mikolov et al. [11], and consists of the two models Continuous Bag-of-Words (CBOW) and Continuous Skip-Gram. The CBOW model aims to predict a word from a window of surrounding words, while the skip-gram architecture uses a current word to predict a window of surrounding words. The GloVe algorithm serves as an extension to the word2vec algorithm, developed by Pennington et al. [10]. GloVe is an unsupervised learning algorithm that uses statistics to produce a word vector space with meaningful substructure. This model outperforms other models on word analogy, word similarity and named entity recognition tasks. In this work, the glove embedding are considered for generating embedding vectors from the input text. In this model, a text such as a sentence or word is represented as the unique integer encoded strings of its corresponding words. The output from this embedding layer will be given as the input to the Deep Learning model to predict the class.

*3. Model Generation*
Deep learning is a sub field of ML that has seen considerable growth in popularity and usefulness in recent years. It will learn complex concepts out of simpler concepts and also that depth enables the computer to learn a multi step computer program. Hence BiRNN is one of the best choice for abusive language detection.

Figure 3 shows the general BiRNN model in tensorflow. Consider the sentences "That son of a bitch insulted my family" and "The bitch won first prize in sporting dogs". One of the problems of RNN is that, to figure out whether the word 'Bitch' is a part of the female dog it's not enough to just look at the first part of the sentence. So to tell, if output should be abusive or normal, need more information than just the first words, because the first words doesn't tell if they'll talking about Bitch as female dog or person. In RNN, all of these blocks are in a forward direction only. So what a BiRNN does is fix this issue. So, a BiRNN works as follows. Consider there are ten inputs or maybe a ten word sentence. This network will have a forward recurrent components and backward recurrent layer. In forward RNN input is passing in the forward direction, which is from the first word to the last. In the backward RNN input sequence is passing in the backward or in reverse direction. So, now this network defines a Acyclic graph. And so, to make the predictions with both the forward activation at time t, and the backward activation at time t being fed to make that prediction at time t. So, if look at the prediction at time t, information from the previous context or the words before 'bitch' are all taken into account with information from the future context or the words after 'bitch'. So this allows the prediction at time t in particular, given a phrase like, "The son of Bitch insulted..." to predict whether Bitch is a part of the person, we take into account information from the past and from the future. So this is the BiRNN and the blocks can also be GRU blocks or LSTM blocks. For a lots of text with NLP problems, a BiRNN with a LSTM appears to be commonly used. Hence this work also used LSTM.

Abusive language detection system proposed here were evaluated using two metrics which are: Accuracy and F1-measure. In order to compute those metrics, we must know about the identifiers which are true positive, true negative, false positive and false negative. True positive (TP) is the text that are correctly classified as abusive, on the other hand true negative (TN) is the text that are correctly classified as normal, however false positive (FP) is the text that are classified as abusive while they are normal and finally false negative (FN) is the misclassified text that are abusive and classified as normal. According to the definition of the four variables TP, TN, FP, FN the metrics will be defined as follows: Accuracy is the percentage of the text that are classified correctly over the total number of text. Furthermore, precision can be defined as how many of classified text are abusive, moreover recall refers to how many of the abusive text are correctly classified as abusive. Another measure that combines precision and recall into one measure is F1-measure. F1-measure and accuracy values must be high in order to get better classification. Figure 5 shows the portion of training BiRNN with attention layer and dropout and having a learning rate of 0.01 with Adam optimizer. The validation accuracy is 0.8007 and the F1 score is 0.8102.

Figure 5. Training summary

## IV. RESULTS AND DISCUSSION

*A. Tools and Libraries*
The tools and libraries used in the proposed system is given in Table 3:

Table 3: Tools used

| Task | Method Used |
|---|---|
| Removal of Unwanted Features | Pandas Library, Regular Expressions |
| Replace repeated characters | Regular Expressions |
| Tokenization | NLTK |
| Conversion to Lowercase | Python String Handling Functions |
| Word segmentation and spelling correction | Ekphrasis |

The sub tasks in the text preprocessing phase has been carried out by Python Pandas library [12], Regular Expressions, Natural Language Tool Kit (NLTK) [13], and some string handling functions in Python.

*B. Results*
Regarding the implemented classification model, several alternatives were considered, from the text preprocessing procedure to model topology. The proposed model is tested by giving various input text from several domains and by also by changing parameters in the network model. The results are also heavily affected by the uneven distribution of data in the target classes. The obtained F1 score is 0.8102 for the proposed model and the accuracy is 0.8007.

In order to get better accuracy the parameters of the deep learning model were tuned and also the obtained result were compared with the existing techniques. Figure 6 and 7 shows testing input results of abusive and hateful classes.

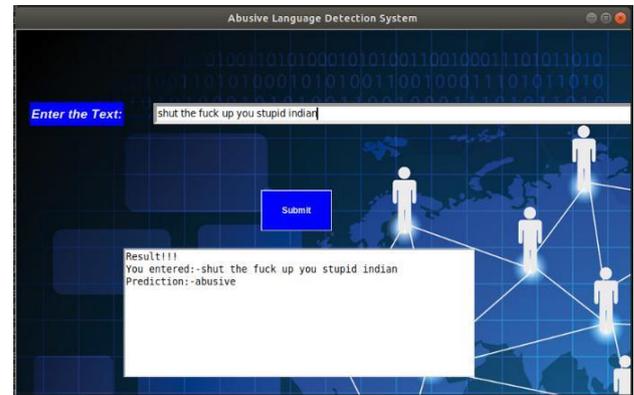

Figure 6. Example for abusive tweet

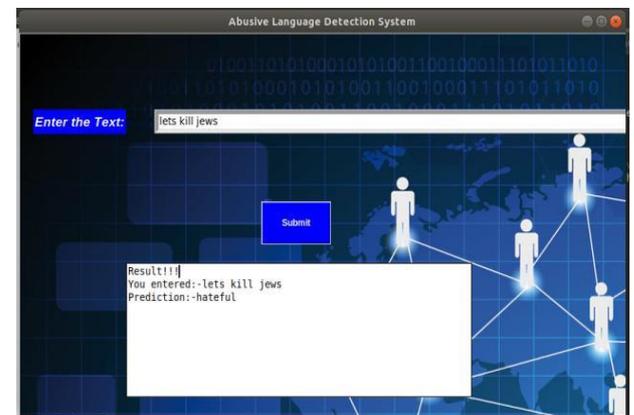

Figure 7. Example for hateful tweet

*C. Performance Analysis*
In deep learning, there are much more parameters that need to be tuned in order to get better accuracy. Testing carried out in different parameters and that choice of optimization algorithm has also impact on the performance. The parameters used for performance analysis are listed below:
    1. Attention mechanism
    2. Learning rate
    3. Optimizer
    4. Dropout

*1. Analysis based on Attention mechanism:*
Many researches proved that the attention mechanism is an effective mechanism to obtain good results in NLP. Here the influence of attention layer [14] in deep learning model on validation accuracy considered. The model is trained with and without attention layer and results were analyzed. The accuracy with attention layer is 80.07% whereas without attention it is not increasing from 65.59% as show in the Table 4. That is attention layer between LSTM layer and output layer helps to focus on important words that have effect on classification. It will assign an attention score to each word in the text. Thus it indicates the amount of attention that the model allocates for each word.

Table 4: Attention analysis

| Parameter | Accuracy (%) |
|---|---|
| BiRNN +Attention | 80.07 |
| BiRNN | 65.59 |

*2. Analysis based on Learning rate:*
The learning rate may be the most important hyper parameter to configure the neural network. It controls how recurrently the model will change in response to the estimated errors in the output of an optimizing problem. Choosing an exact learning rate is challenging task. When its value is too small may result in a long training process, while a value which is too large may result in sub-optimal solutions, i.e., the neural network converges too fast to an unfeasible solutions. Thus it is vital to know how to investigate the effects of learning rates on model performance and to build an intuition on the dynamics of learning rates on model behavior. The impact of learning rate in accuracy is shown in Table 5.

For learning rate of 0.001 the accuracy was 79.91% but for 0.01 it was 80.07%. Since the accuracy reduced after that, for the experimental purpose we fixed learning rate fixed as 0.01to make better prediction with deep learning model.

Table 5: Impact of Learning rate

| Learning rate | Accuracy (%) |
|---|---|
| 0.001 | 79.30 |
| 0.005 | 79.91 |
| 0.01 | 80.07 |
| 0.05 | 61.19 |
| 0.1 | 55.89 |

*3. Analysis based on Optimizer:*
Optimizers are algorithms or methods used to change the attributes of the neural network, such as weights and learning rates, in order to reduce losses. How we should change the weights or learning rates of your neural network to reduce losses is defined by the optimiser that we use. Optimization algorithms or strategies are able to reduce losses and delivering the most accurate results possible. Here the performance of Gradient Descend and Adam optimizer were analysed.

Gradient Descent is the most basic and widely used optimization algorithm. It is a first-order optimization algorithm and dependents on the derivative of first-order loss function. It calculates how the weights should be changed such that the function can reach a minimum. Through back propagation, the loss is transferred from one layer to another and the weights are modified depending on it. Adaptive Moment Estimation (Adam) is the best optimiser to train the neural network in less time and more efficiently [15]. It works with first-and second-order momentum. Since we want slow and careful search, Adam is chosen as our optimizer.

Table 6: Optimizer analysis

| Parameter | Accuracy (%) |
|---|---|
| Adam | 80.07 |
| Gradient descend | 69.95 |

In the proposed work Adam optimizer provided better results than Gradient Descend optimizer. The accuracy was 80.07% for Adam and it is only 69.95% for Gradient Descent shown in Table 6.

*4. Analysis based on Dropout:*
Deep learning neural networks may quickly over-fit a dataset of training with few examples. Sets of neural networks with different model configurations are known to reduce over-fitting, but require additional computational training and maintenance costs for multiple models. A single method that can be used is dropping out of nodes during training. This is called dropout [16] and offers a very computationally cheap and highly effective regularization method to reduce over-fitting and improve generalization error in deep neural networks of all kinds.

By removing this dropout layer only, the model produce an accuracy of 78.06% as shown in Table 7 which is low compared to the model with dropout.

Table 7: Dropout analysis

| Parameter | Accuracy (%) |
|---|---|
| With dropout | 80.07 |
| Without Dropout | 78.06 |

*D. Performance Comparison*
The comparison between the proposed system and the existing methods [7] which had used CNN and RNN for abusive language detection is given in Table 8.

Table 8: Comparison

| Model | F1 Score |
|---|---|
| CNN | 0.7840 |
| RNN | 0.8010 |
| BiRNN | 0.8102 |

It shows the F1 score of CNN, RNN, and BiRNN for the same multi labelled dataset. When compared to the performance of CNN and RNN, BiRNN gives best performance based on the dataset.

From this result it is clear that BiRNN is a better deep learning model compared to other models when we are considering the various abusive behaviors in Twitter platform.

### V. CONCLUSION AND FUTURE SCOPE

The proposed work presented the use of BiRNN that can focus on various abusive behaviors on Twitter platform. Also by comparing the model with CNN and RNN, the work proved that the proposed BiRNN is a better deep learning model when we are considering the multiple abuse related labels in social media. The different experimental results show that there is a significant performance improvement using the proposed model.

In future, the performance of proposed system shall be investigated on several domain datasets like Facebook, Wikipedia, Twitter, etc., in order to generalize the behavior of users of several online communities.

## AUTHORS PROFILE

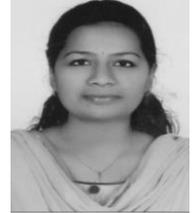

**Dincy Davis** pursed Bachelor of Technology (B. Tech) in Computer Science from College of Engineering Kottarakara, Kollam, India in 2016. She is currently pursuing Master of Technology (M. Tech) in Computer Science from Government Engineering College Palakkad, India. She is specialized in Computational Linguistics and area of interests includes Machine Learning, Natural Language Processing, Artificial Intelligence, Data Mining, Speech Processing, and Information Retrieval. She had done 5 months Machine Learning Internship program at Ignitarium, Kochi infopark, Kerala and has 1 years of Graduate Apprentice experience from Kerala State IT Mission.

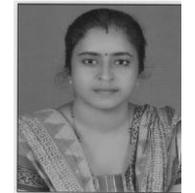

**Reena Murali** pursed B.Tech from Mangalore University in 1996, M.Tech from IIT Madras in 2006, Ph.D from CUSAT in 2016. She is currently working as Professor in Department of Computer Science and Engineering at Government Engineering College Palakkad. She has published many research papers in reputed international conferences and journals including GENE, Elsevier. Her main research work focuses on Bioinformatics. She has 20 years of teaching experience.

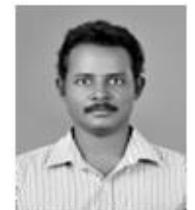

**K R Remesh Babu** received Ph.D Degree from Cochin University of Science and Technology (CUSAT) in 2019. He also holds bachelor degree in Mathematics, Information Technology and master degree in Computer Science. Currently he is working as Associate Professor and Head in the department of Information Technology, Government Engineering College, Palakkad. He is interested in Distributed and Cloud Computing, Internet of Things, Wireless Sensor Networks, Machine Learning, and Big Data Analytics.